\pdfoutput=1

\documentclass[11pt]{article}

\usepackage{ACL2023}

\usepackage{times}
\usepackage{latexsym}

\usepackage[T1]{fontenc}

\usepackage[utf8]{inputenc}

\usepackage{microtype}

\usepackage{inconsolata}

\usepackage{epsfig}
\usepackage{graphicx}
\usepackage{amsmath}
\usepackage{amssymb}
\usepackage{xspace}
\usepackage{amstext}
\usepackage{multirow}
\usepackage{booktabs}   
\usepackage{subcaption}
\usepackage{colortbl}
\usepackage{hyperref}
\usepackage{todonotes}
\usepackage{placeins}
\usepackage{wrapfig}
\usepackage{longtable}
\usepackage{listings}

\newcommand{\rparagraph}[1]{\vspace{1.4mm}\noindent\textbf{#1.}}

\newcommand{\sparagraph}[1]{\vspace{0.0mm}\noindent\textbf{#1.}}

\newcommand{\rparagraphnodot}[1]{\vspace{1.6mm}\noindent\textbf{#1}}

\newcommand{\bench}{\texttt{FOCI}\xspace}

\title{African or European Swallow? Benchmarking Large Vision-Language Models for Fine-Grained Object Classification}

\author{\textbf{Gregor Geigle$^{12}$ \quad \quad Radu Timofte$^{2}$ \quad \quad Goran Glava\v{s}$^{1}$} \\
  $^{1}$W{\"u}NLP, $^{2}$Computer Vision Lab, CAIDAS, University of W{\"u}rzburg,  \\
  \texttt{gregor.geigle@uni-wuerburg.de}}
  
\begin{document}
\maketitle
\begin{abstract}
%
%

Recent Large Vision-Language Models (LVLMs) demonstrate impressive abilities on numerous image understanding and reasoning tasks.
The task of fine-grained object classification (e.g., distinction between
\textit{animal species}), however, has been probed insufficiently, despite its downstream importance.
We fill this evaluation gap by creating
\texttt{FOCI}\xspace (\textbf{F}ine-grained \textbf{O}bject \textbf{C}lass\textbf{I}fication),
a difficult multiple-choice benchmark for fine-grained object classification, from existing object classification datasets: 
(1) 
multiple-choice avoids ambiguous answers associated with casting classification as open-ended QA task; 
(2) 
we retain classification difficulty by mining negative labels with a CLIP model. 
\texttt{FOCI}\xspace
complements five popular classification datasets with four domain-specific subsets
from ImageNet-21k. 
We benchmark 12 public LVLMs on \texttt{FOCI}\xspace and show that it tests for a \textit{complementary skill} to established image understanding and reasoning benchmarks. 
Crucially, CLIP models exhibit dramatically better performance 
than LVLMs. Since the image encoders of LVLMs come from these CLIP models, this points 
to inadequate alignment for fine-grained object distinction between the encoder and the LLM 
and warrants (pre)training data with more fine-grained annotation.
We release our code at \url{https://github.com/gregor-ge/FOCI-Benchmark}.

\end{abstract}


\section{Introduction}

Large Vision Language Models (LVLMs)---Large Language Model (LLM) that have been adapted to process images as input alongside text---have shown impressive performance on a wide range vision-language tasks  \cite{li_blip-2_2023,liu_visual_2023,openai_gpt-4_2023,anil_gemini_2023}. 
LVLMs are mutually compared using a range of benchmarks that test for various image understanding and reasoning skills, such as existence and counting of objects, localization, comparison between objects or identifying object attributes \cite{goyal_making_2017,hudson_gqa_2019,liu_mmbench_2023}.

\begin{figure}[t]
    \centering
    \includegraphics[width=0.99\linewidth]{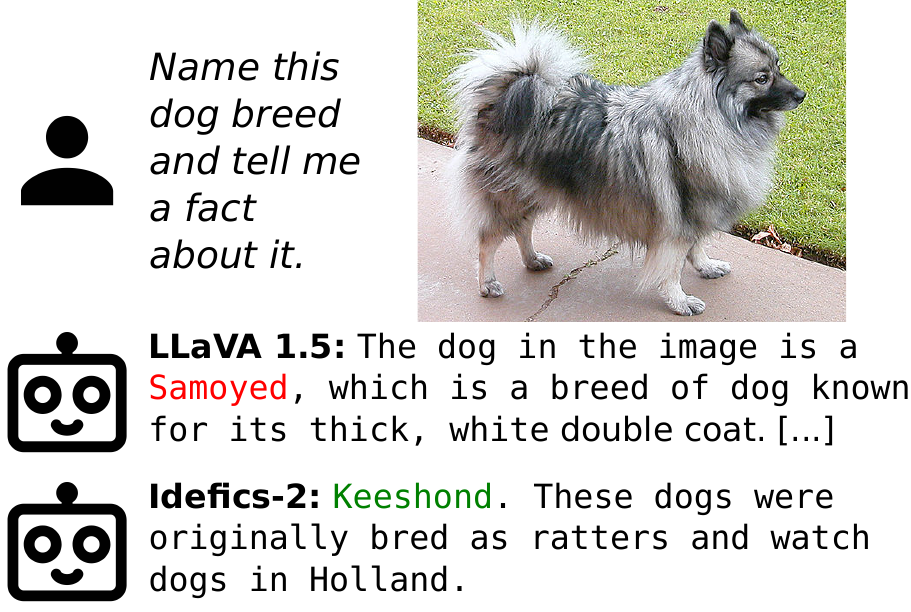}
    \caption{The importance of object recognition: LLaVA 1.5 fails to identify the dog breed. Idefics-2 correctly recognizes it and gives a correct fact as a result.}
    \label{fig:dog_example}
    \vspace{-1em}
\end{figure}

LVLMs are, however, barely ever tested for fine-grained object classification---the ability to correctly recognize different animals, plants, or man-made objects---which is, we argue, an important skill that complements general image understanding.\footnote{For simplicity, we use `object' to refer to both living entities like animals as well as to inanimate objects.} Besides it being an end-task in itself, e.g., to answer questions such as ``\textit{What is this flower called?''}, it is often implicitly needed in information-seeking situations, where the success depends on the models' ability to correctly and precisely identify an object \cite{hu_open-domain_2023,chen_can_2023,mensink_encyclopedic_2023}. 
As illustrated in Figure~\ref{fig:dog_example}, only one of the LVLMs correctly identifies the dog breed in the image and can follow up with relevant information. Note that this is different from general L(V)LM hallucination \cite{zhang_sirens_2023}, where models `invent' 
incorrect information. Instead, the generated content is correct for the object, but the object is misclassified: 
the information about \textit{Samoyed} by LLaVa 1.5 is correct, but the dog in the image is a \textit{Keeshond}.

To fill this gap in LVLM evaluation, we create a comprehensive benchmark dubbed \bench (\textbf{F}ine-grained \textbf{O}bject \textbf{C}lass\textbf{I}fication) that tests models' fine-grained object recognition over a wide range of object categories. 
Our key contribution is a well-defined task formulation that avoids pitfalls of prior work:
We argue that an open question answering (QA) formulation (i.e., answer the question ``What is this''?), as done, e.g., by \citet{xu_lvlm-ehub_2023}, is an ill-defined task for two reasons. \textbf{1)} the \textit{complete} set of admissible answers is \textit{not} provided (e.g., admissible answers for the dog in Figure~\ref{fig:dog_example} include \textit{Keeshond}, \textit{Dutch Barge Dog}, 
and \textit{Wolfspitz}).
For objects with only a few synonym labels, one can provide all answer options but this does not scale to hundreds or thousands of objects. Constrained decoding to only the admissible labels is computationally expensive for large label sets \cite{chen_pali_2022}. 
\textbf{2)} The expected taxonomy level of the answer is not specified.
For the given example, \textit{dog}, \textit{Spitz}, and \textit{Keeshond} are all ontologically correct answers; but recognizing a \textit{Keeshond} is much more difficult than recognizing a \textit{dog}.     
To address the above shortcomings, we formulate object classification as a multiple-choice problem
To avoid that the reduction to only a handful candidate answers renders the task trivial, we use a CLIP model \cite{radford_learning_2021} in a zero-shot configuration to mine difficult choices from the pool of class labels. 
We assemble \bench from 5 popular classification datasets for different domains (flowers, cars, food, aircraft, pets)
and additionally create 4 domain subsets from ImageNet-21k \cite{deng_imagenet_2009} for \textit{animals}, \textit{plants}, \textit{food}, and \textit{man-made} objects. 

We extensively evaluate 12 publicly available LVLMs on \bench and find that many of them like the popular LLaVA 1.5 struggle with fine-grained object classification. We observe that models with similar performance on established benchmarks can yield quite different and uncorrelated results on \bench, highlighting that fine-grained object classification is indeed a \textit{distinct skill} for LVLMs, and that \bench should thus complement existing image understanding and reasoning benchmarks. 
Comparing the models further, we observe that the scale of their (pre-)training data seems to impact their performance on \bench significantly more than for image understanding tasks.
A comparison with the underlying CLIP models used as the LVLMs' image encoders shows that the encoder's zero-shot accuracy provides an upper bound for the LVLM, with the LVLM performance lagging drastically behind. This suggests that the alignment between the image encoder and LLM in LVLMs seems to be insufficiently semantically fine-grained. 
%
We finally perform controlled experiments to isolate the modeling and training decisions that impact the models' performance in \bench. As is the case with other benchmarks, both larger LLMs and stronger image encoders improve results. Most importantly, incorporating captions into the training data that explicitly name the downstream objects helps with classification. Similarly, including fine-grained classification objectives to the training mix can improve models' \bench performance.


\begin{figure*}
    \centering
    \includegraphics[width=0.9\linewidth]{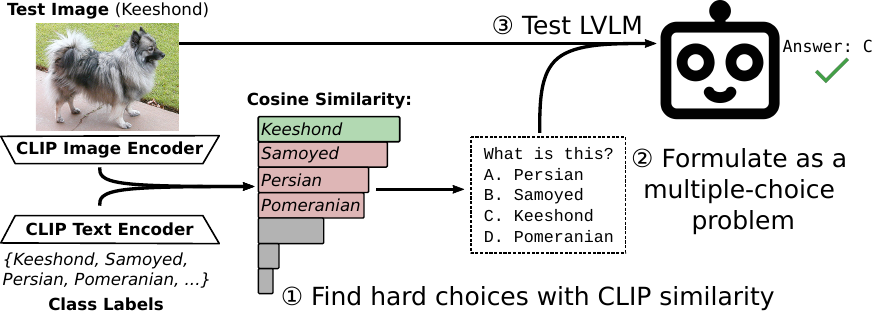}
    \caption{Testing LVLMs on object classification through multiple-choice: (1) We compute the CLIP cosine similarity between a test image and class labels; we select the correct label and the three most similar (wrong) labels to (2) formulate a multiple-choice problem, which (3) is given to the LVLM who has to predict the correct choice.
    }
    \label{fig:creation_process}
    \vspace{-1em}
\end{figure*}

\section{Related Work}
\label{sec:rw}

\sparagraph{Large Vision-Language Models}
LVLMs align pre-trained image encoders (generally a Vision Transformer (ViT) \cite{dosovitskiy_image_2021} from CLIP  \cite{radford_learning_2021}) to a Large Language Model (LLM), yielding an LLM that can work with images as input besides text \cite{chen_pali_2022,alayrac_flamingo_2022,li_blip-2_2023,dai_instructblip_2023,liu_visual_2023,liu_improved_2023,bai_qwen-vl_2023,laurencon_obelisc_2023,chu_mobilevlm_2023,zhang_internlm-xcomposer_2023}.
LVLMs are commonly trained in two stages: first, an alignment module 
between the image encoder and the LLM---a shallow feed-forward network \cite{liu_visual_2023,liu_improved_2023} or more complex modules like a resampler \cite{alayrac_flamingo_2022,li_blip-2_2023}---that projects image tokens into the LLM input embedding space is trained using image-caption pairs. In the second stage, the model is trained for general-purpose inference on a mix of tasks, e.g., visual Q\&A \cite{goyal_making_2017,hudson_gqa_2019} and (visual) chat instruction data \cite{chiang_vicuna_2023,liu_visual_2023}.
While the second stage is fairly similar across the recent models, the first stage is where training greatly varies: on the low end, models are trained with less than a million examples \cite{liu_improved_2023,liu_llava-next_2024}; on the high end, over a billion image-text pairs are used \cite{bai_qwen-vl_2023,dong_internlm-xcomposer2_2024,laurencon_what_2024}. Despite differences in data size,
 models on both ends of the spectrum can achieve competitive results on popular benchmarks.
In this work, we show that better visio-linguistic alignment in the first training stage substantially boosts fine-grained object classification abilities.

\rparagraph{Benchmarking LVLMs}
Most existing benchmarks, e.g., VQAv2 \cite{goyal_making_2017}, GQA \cite{hudson_gqa_2019}, MME \cite{fu_mme_2023}, MMBench \cite{liu_mmbench_2023}, or Seed-Bench \cite{li_seed-bench_2023}, test LVLMs for image understanding and reasoning capabilities such as recognition of color and other attributes, object counting, recognizing object position and orientation and similar. Other benchmarks like MMMU \cite{yue_mmmu_2023} test world knowledge and reasoning capabilities in different domains.
Although (fine-grained) object classification is a prominent end-task in itself and relevant in conversational applications, it is barely considered in LVLM evaluation protocols. 
The work that addresses the task is limited. \textit{(i)} Models with in-context learning capabilities are evaluated on few-shot object classification but the models do not classify images in isolation and instead compare the target image with labeled in-context examples \cite{tsimpoukelli_multimodal_2021,alayrac_flamingo_2022}. \textit{(ii)} Pali \cite{chen_pali_2022} was evaluated on ImageNet \cite{deng_imagenet_2009} by scoring every class labels, which is computationally expensive. \textit{(iii)} LVLM-e-Hub \cite{xu_lvlm-ehub_2023} includes some image classification datasets (like ImageNet) but they formulate it as open-ended QA task with ambiguity over expected answers, which leads to low accuracy scores for all models. \textit{(iv)} In knowledge-intensive VQA, models have to recognize the correct object 
(e.g., a specific building) to answer correctly; objects are recognized either implicitly (the QA model needs to know which object it is to answer correctly) or explicitly when a knowledge base is used to retrieve relevant information \cite{hu_open-domain_2023,chen_can_2023,mensink_encyclopedic_2023}.
In contrast to these efforts, we propose a standardized evaluation of LVLMs for (fine-grained) object classification by converting image classification datasets into difficult multi-choice tasks.

\section{Multiple-Choice Image Classification}

Image classification is a fundamental problem in computer vision with a plethora of datasets available. 
In this work, we focus on \textit{fine-grained object classification} where models have to differentiate between several objects belonging to a specific domain, e.g., animal species or car models.
%
We leverage existing datasets as resources for annotated data and frame object classification as a multiple-choice task with well-defined answer candidates.

\rparagraphnodot{Why Multiple-Choice?} The standard formulation of object classification tasks for LVLMs is via question answering, with open-ended answer generation \citet{xu_lvlm-ehub_2023-1}. This formulation, we argue, represents an ill-posed problem for two main reasons: (1) the expected level of granularity in the object taxonomy that is expected as the answer is not defined, and is difficult to define in general (e.g., for the image from Figure \ref{fig:dog_example}, \textit{dog}, \textit{Spitz}, or \textit{Keeshond} are all correct labels); (2) the set of admissible answers in existing datasets is \textit{not complete}: most objects have multiple synonymous labels, all of which constitute a correct answer (e.g., \textit{Keeshond}, \textit{Dutch Barge Dog}, and \textit{Wolfspitz}), but only subsets of those are provided as admissible labeles in existing datasets. 
Providing complete synonym sets and specifying the expected level of granularity of the answer is, in the general case, infeasible. 
Instead, we propose to formulate fine-grained object classification as a multi-choice task, where the models are provided with a set of candidate answers from which the correct answer is to be selected; this way the expected (i.e., correct) output is well-defined. 

\rparagraph{Mining Hard Choices} To maintain difficulty despite the reduction to only a small set of candidate labels, we mine for each example image \textit{difficult} incorrect labels from all class labels used in the concrete image classification dataset. We argue that a reduction to the most likely incorrect classes retains the task difficulty as even in classification over large class sets (e.g., thousands of classes), models easily discern between unrelated classes and most errors stem from close classes anyways (e.g., in the Oxford-Pets dataset, which covers 37 cat and dog breeds, cat breeds are irrelevant for dog images).
We use a CLIP model for mining difficult candidates: for every example image, we select the three most similar (incorrect) class labels as negative choices. We rank the dataset classes for an image using the standard CLIP zero-shot setup: the text encoder embeds all class labels, the image encoder embeds the image, and the class labels are ranked in decreasing order of cosine similarity of their respective text embeddings with the image embedding. 
We avoid biasing the choice selection towards any concrete LVLM in our evaluation by selecting \texttt{OpenCLIP ViT-L/14} \cite{ilharco_openclip_2021}: 
its image encoder has not been used by any of the LVLMs. 
Figure~\ref{fig:creation_process} illustrates both the process of mining negatives for an image and testing an LVLM on the resulting set of candidate choices.





\rparagraph{\bench (\textbf{F}ine-grained \textbf{O}bject \textbf{C}lass\textbf{I}fication)}
%
%
%
We collate our \bench benchmark from diverse existing datasets, selecting in all cases four candidate choices for each image (i.e., the correct label and three most similar negatives). We complement (1) established datasets commonly used for evaluating CLIP models \cite{radford_learning_2021,ilharco_openclip_2021} with (2) additional challenging larger-scale datasets that we derive from ImageNet-21k \cite{deng_imagenet_2009}. 
For the former, we select the following five datasets:
\textbf{FGVC-Aircraft} \cite{maji_fine-grained_2013} contains images of 100 different aircraft types;
\textbf{Flowers102} \cite{nilsback_automated_2008} contains images of 102 different flower species;
\textbf{Food101} \cite{bossard_food-101_2014} covers 101 dishes;
\textbf{Oxford-Pet} \cite{parkhi_cats_2012} contains images of 37 cat and dog breeds.
\textbf{Stanford-Cars} \cite{krause_3d_2013} covers 196 car models.

As some of the above datasets are not particularly challenging for existing CLIP models in zero-shot evaluations, we additionally construct four new challenging datasets from ImageNet-21k (\textbf{IN-21k)}. We first merge ImageNet-COG \cite{sariyildiz_concept_2021} (5k classes) and ImageNet-1k (\textbf{IN-1k}), for a total of 6k classes that are all leaf nodes in the WordNet \cite{miller_wordnet_1994} taxonomy: this means that no two labels stand in the \textit{is-a} 
relation and there cannot be multiple correct answers stemming from different taxonomy levels (e.g., \textit{dog} and \textit{Pomeranian}). Next, we group the classes according to their WordNet lexicographer file names, and create a dataset for each of the four most represented ones: \textbf{Animal} (1322 classes), \textbf{Plant} (957 classes), \textbf{Food} (563 classes), and \textbf{Artifact} (man-made objects, 2631 classes).
We prepend \textbf{IN-} (ImageNet-) in our experiment to mark these datasets.

One could, in principle, add more object types and domains to the evaluation: our goal was to include a reasonably diverse set of domains, from which, when put together in a benchmark, one could reliably extrapolate general fine-grained object recognition abilities of LVLMs.    
For further analysis, in Appendix~\ref{sec:appendix:more_eval} we additionally evaluate LVLMs on more general (i.e., not domain-specific) object classification under different image distribution shifts (using ImageNet-1k) and for geographic distribution shifts with common objects photographed in different regions of the world, using GeoDE \cite{ramaswamy_geode_2023}.

\section{Evaluating Public LVLMs}

We evaluate 12 diverse and publicly available LVLMs on \bench.
We then analyze how the performance of LVLMs relates to the results of their underlying CLIP image encoders.

\begin{table}[t]
    \centering
     \def\arraystretch{0.97}
     \resizebox{1\linewidth}{!}{
    \begin{tabular}{lrrr}
    \toprule
    \bf Model & \bf \#P & \bf Pretrain & \bf Task Mix \\
    \midrule
    Idefics-1 \cite{laurencon_obelisc_2023} & 9B &  350M & 1M \\
    Idefics-2 \cite{laurencon_what_2024} & 8B & 1.5B & ? \\
    BLIP2 Flan-T5-XL \cite{li_blip-2_2023} & 4B & 130M & --- \\
    InstructBLIP Flan-T5-XL \cite{dai_instructblip_2023} & 4B & 130M & 1M \\
    InstructBLIP Vicuna \cite{dai_instructblip_2023} & 8B & 130M & 1M \\
    InternLM XComposer 2 \cite{dong_internlm-xcomposer2_2024} & 7B & >1B & 600M \\
    LLaVA 1.5 \cite{liu_improved_2023} & 7B & 560k & 660k \\
    LLaVA-Next (Mistral) \cite{liu_llava-next_2024} & 7B & 560k & 760k \\
    MobileVLM V2 \cite{chu_mobilevlm_2024} & 7B & 1.2M & 2.4M \\
    Pali-Gemma$^1$ & 3B & >1M & ? \\
    Phi-3-Vision \cite{abdin_phi-3_2024} & 4B & >10M & >1M \\
    Qwen-VL-Chat \cite{bai_qwen-vl_2023} & 10B & 1.4B & 50M \\
    \bottomrule
    \end{tabular}
    }
    \caption{The 12 tested public LVLMs. We provide parameters count (\#P; LLM + image encoder parameters) and the dataset size (in images) used during the pretraining and task mix training phase. For some fields, we put a conservative estimate or `?' if no estimate is possible. 
    $^1$\href{https://ai.google.dev/gemma/docs/paligemma/model-card}{Model Card}, tech report pending at time of writing.
    }
    \label{tab:model_infos}
\end{table}

\rparagraph{Model and Inference Details.}
Our selected models span a variety of architectures and training paradigms. Table~\ref{tab:model_infos} summarizes key information (the number of parameters and the size of the training data) for each model.
Due to our hardware constraints, we benchmark models with LLMs having $\leq$7B parameters.
At inference time, we provide the LVLMs with the image and the four candidate choices. The choices are in random order to avoid model-specific preferences for answer positions \cite{liu_mmbench_2023}); the model provides as output one of the choices, which is compared with the ground truth label: we then report the performance in terms of accuracy.
See Appendix~\ref{sec:appendix:eval_details} for further details on models, the inference setup, and datasets.


\subsection{Results}

\begin{table*}[t]
    \centering
     \def\arraystretch{0.97}
     \resizebox{0.99\linewidth}{!}{
    \begin{tabular}{lrrrrrrrrrr}
    \toprule
\bf Model & \bf IN-Food & \bf IN-Artifact & \bf IN-Animal & \bf IN-Plant & \bf Aircraft & \bf Flowers102 & \bf Food101 & \bf O.-Pet & \bf S.-Cars & \bf $\varnothing$ \\
\midrule 
Idefics-1               & 40.18 & 41.90 & 31.37 & 29.55 & 34.62 & 51.70 & 72.44 & 48.51 & 29.42 & 42.19 \\
Idefics-2               & \textbf{56.38} & \textbf{52.56} & 46.50 & \textbf{41.47} & \textbf{56.23} & 72.78 & \textbf{89.70} & 81.28 & \textbf{80.25} & \textbf{64.13} \\
BLIP-2 Flan-T5-XL       & 51.47 & 47.41 & 39.22 & 32.59 & 32.94 & 64.32 & 82.51 & 65.00 & 67.68 & 53.68 \\
InstructBLIP Flan-T5-XL & 49.25 & 47.83 & 38.07 & 32.88 & 29.19 & 62.29 & 76.77 & 59.99 & 64.58 & 51.21 \\
InstructBLIP Vicuna     & 43.94 & 42.39 & 37.32 & 30.04 & 31.68 & 50.90 & 63.47 & 54.92 & 48.25 & 44.77 \\
InternLM XComposer 2    & 50.43 & 47.84 & 38.98 & 33.23 & 40.53 & 54.25 & 79.30 & 63.23 & 53.89 & 51.30 \\
LLaVA 1.5               & 47.76 & 45.61 & 36.32 & 33.00 & 34.71 & 51.37 & 72.80 & 52.25 & 46.92 & 46.75 \\
LLaVA-Next              & 46.32 & 45.54 & 35.51 & 31.86 & 32.49 & 43.91 & 71.30 & 53.72 & 49.48 & 45.57 \\
MobileVLM v2            & 46.50 & 44.58 & 37.60 & 33.75 & 35.01 & 54.89 & 74.38 & 53.69 & 46.29 & 47.41 \\
Pali-Gemma              & 54.25 & 48.79 & 42.28 & 37.04 & 39.87 & 69.64 & 82.36 & 75.42 & 64.64 & 57.14 \\
Phi-3-Vision            & 46.66 & 42.75 & 35.11 & 31.27 & 42.33 & 51.59 & 69.98 & 56.36 & 54.50 & 47.84 \\
Qwen-VL-Chat            & 52.36 & 50.95 & \textbf{48.45} & 40.09 & 45.96 & \textbf{75.95} & 83.92 & \textbf{87.82} & 76.23 & 62.41 \\
\bottomrule
    \end{tabular}
    }
    \caption{
    Accuracy on \bench: on individual datasets and average ($\varnothing$), for the 12 tested public LVLMs.
    }
    \label{tbl:icbench}
    \vspace{-1em}
\end{table*}

\sparagraph{\bench vs. Other Benchmarks} 
Table~\ref{tbl:icbench} displays the results for the 12 benchmarked LVLMs on \bench. We first compare the models' performance and relative ranking on \bench with their results on popular image understanding benchmarks (we show the models' performance on GQA \cite{hudson_gqa_2019}, MMBench \cite{liu_mmbench_2023}, and MMMU \cite{yue_mmmu_2023} in Table~\ref{tab:appendix:standard_benchmarks} in the Appendix~\ref{sec:appendix:general_eval}). Model's results on \bench are much less correlated with their respective results on other benchmarks:
better results on GQA, MMBench, or MMMU do not necessarily imply better results for fine-grained object classification and vice versa.
Qwen-VL, for example, is amongst the best-performing models in object classification in \bench, but is fares much worse on the standard benchmarks, where several yield better results. On the other hand, Phi-3-Vision has among the best results on the standard benchmarks but exhibits only average performance on \bench.
These results indicate that fine-grained object classification is a skill that is complementary to what other image understanding and reasoning benchmarks test and as such should be added to LVLM evaluation protocols. 

\rparagraph{Training Data} 
One important factor for strong object recognition on \bench seems to be the amount of image-text data used for (pre-)training the alignment component of the LVLM in the first training phase (see \S\ref{sec:rw}). 
On the common understanding benchmarks, models like LLaVA 1.5, and LLaVA-Next show strong results despite being pretrained with $<$1M image-text pairs.
However, the two best models on \bench, Idefics-2 and Qwen-VL, are both pretrained on $\sim$1.5B images and drastically outperform the LLaVA models. This suggests that object classification requires larger-scale training for a much more fine-grained alignment between the image encoder and LLM, compared to what is needed in general for image understanding. We isolate the effect of the alignment training data (in a smaller-scale setup) in \S\ref{sec:experiments}.
The results for InstructBLIP are somewhat inconclusive: with Flan-T5-XL as LLM, it exhibits good \bench performance, but with Vicuna (and otherwise identical training) the results are substantially worse. This would suggest that, other than the scale of the alignment training, the LLM itself plays an important role.

\rparagraph{Other Factors} 
Very high image resolution, which is highly beneficial for OCR-heavy tasks like chart understanding \cite{liu_llava-next_2024}, does not seem to be relevant for fine-grained object classification. This stems from the comparison between LLaVA 1.5 and LLaVA-Next,  where the latter's main difference w.r.t. the former is training with (and inference on) images of higher resolution. This is unsurprising as images in object classification datasets typically contain large centered objects, making larger resolution unnecessary for solving the task.
The LLM and image encoder are likely also major factors for the ultimate performance but we cannot isolate them in this observational analysis; instead, we consider them in controlled experiments in \S\ref{sec:experiments}.

\subsection{LVLM vs. Its Corresponding CLIP}

\begin{figure}[t]
    \centering
    \includegraphics[width=0.9\linewidth]{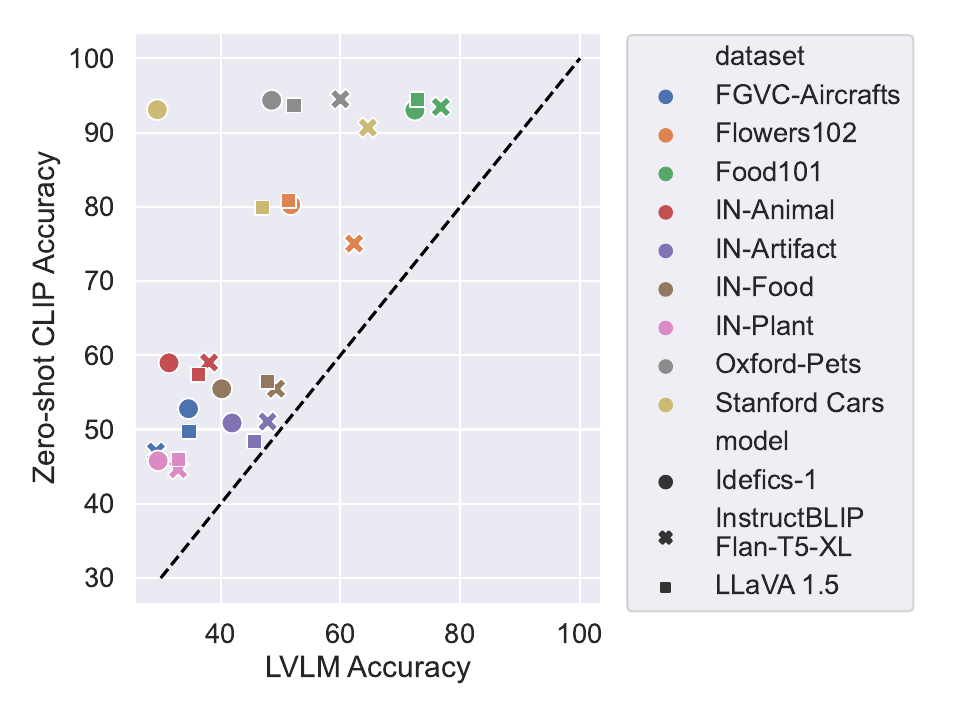}
  \caption{
We plot the LVLM accuracy against the CLIP zero-shot accuracy (using the 4 multiple-choice options for CLIP for a fair comparison) of the underlying CLIP image encoder used by the LVLM.
  }
\label{fig:plot:clip_lvlm_scatter}
\vspace{-1em}
\end{figure}

\begin{figure}[t]
    \centering
    \includegraphics[width=0.9\linewidth]{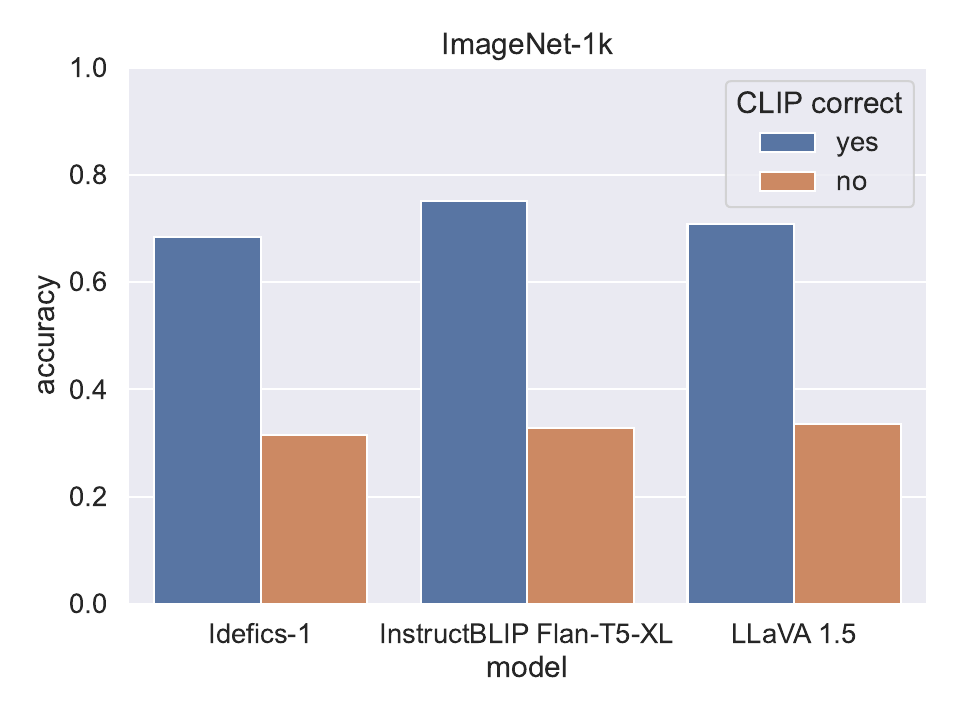}
  \caption{
  Accuracy of three LVLMs on ImageNet-1k, for example subsets on which the zero-shot classification with the corresponding CLIP model is (in)correct.
  }
\label{fig:plot:clip_wrong_imagenet}
\vspace{-1em}
\end{figure}

Several of the tested LVLMs keep their underlying CLIP image encoder frozen throughout training. This means that the cross-modal alignment between the CLIP's image encoder and its text encoder is untouched, allowing us to compare the performance of these LVLMs directly against the CLIP models from which they take the image encoder.   

Specifically, we consider three LVLMs with their corresponding CLIP models: Idefics-1, which uses \texttt{OpenCLIP ViT-H/14} \cite{ilharco_openclip_2021}, LLaVA 1.5, which uses \texttt{OpenAI ViT-L/14} \cite{radford_learning_2021-1}, and InstructBLIP Flan-T5 with \texttt{EVA-1 ViT-g/14} \cite{fang_eva_2022-1}.


\rparagraphnodot{CLIP Zero-Shot Classification as Upper Bound.} 
The image and text encoder of a CLIP model were trained jointly on huge datasets;
in contrast, the alignment of the CLIP's image encoder to the LLM is learned with comparatively less image-text data (e.g., InstructBLIP is pre-trained with 100M samples while EVA-1 was trained with 11B samples).
We compare in Figure~\ref{fig:plot:clip_lvlm_scatter} the LVLM performance against the zero-shot classification accuracy of the corresponding CLIP model (for a fair comparison, CLIP only considers the same 4 labels as LVLM does in multiple-choice formulation). We observe that the LVLM performance is indeed consistently lower than that of the corresponding CLIP model.
However, while the CLIP zero-shot classification accuracy seems to be an upper bound for the LVLM, the gaps vary substantially across the \bench datasets: from <10\% on IN-Artifact to 40-50\% on Oxford-Pets.
These results indicate that, while the alignment between the image encoder and LLM is undertrained in general, there are also drastic differences in the quality of alignment for different types of objects (i.e., domains). For certain domains (e.g., Oxford-Pets) the LLM seems to struggle to process the image features, despite the CLIP image encoder encoding sufficient information (as evidenced by the much better corresponding CLIP performance). 

\rparagraphnodot{CLIP wrong$\implies$LVLM wrong?}
We analyze the predictions of LVLMs on instances that the corresponding CLIP model misclassifies to measure whether those classification errors  propagate to the LVLM: in other words, if the CLIP model is wrong, is the LVLM using its image encoder also bound to misclassify the image? 
Figure~\ref{fig:plot:clip_wrong_imagenet} summarizes the results of this analysis on ImageNet-1k (in our multi-choice formulation) for three LVLMs; for the \bench datasets we provide the same analysis in Figure~\ref{fig:appendix:clip_error} in the Appendix.
We observe that LVLM accuracy plummets on examples on which the corresponding CLIP fails: in fact, for instances that CLIP cannot correctly classify, the performance of the corresponding LVLM gets close to random (25\%) for all three LVLMs in the analysis. 
These observations---that CLIP performance is an upper-bound for LVLM accuracy and that its errors propagate to the LVLM---highlight that the selection of an image encoder is a key design decision for LVLMs performance and suggest that future improvements in image encoding are likely to also propagate to LVLM object recognition capabilities.

\section{Controlled Experiments}
\label{sec:experiments}

We next perform a set of controlled experiments to disentangle the effects of individual LVLM design choices on (fine-grained) object classification.
Our analysis encompasses three main factors: (1) the LLM size, (2) the image encoder, and (3) targeted changes to the training data. For (2) and (3), we train LVLMs following the LLaVA 1.5 recipe with \texttt{StableLM 2 Zephyr 1.6B} \cite{bellagente_stable_2024} as LLM and \texttt{OpenAI CLIP-L/14-224} as the image encoder (see the Appendix~\ref{sec:appendix:train_details} for training details).

\begin{figure}[t]
    \centering
    \includegraphics[width=0.9\linewidth]{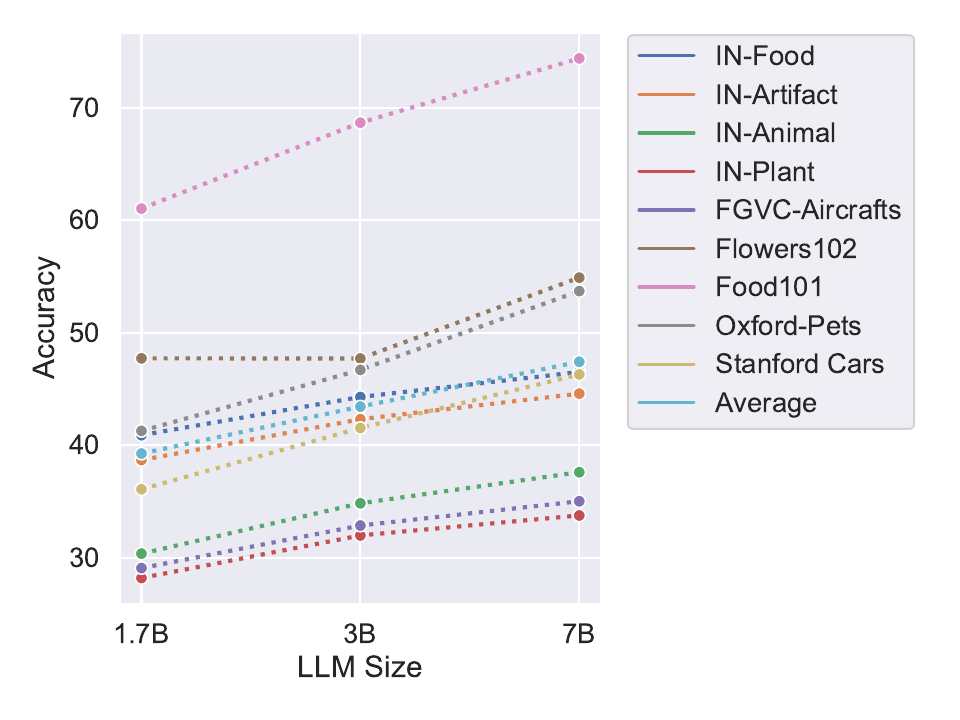}
    \caption{Results with MobileVLM v2 over its three LLM sizes with otherwise identical training.}
    \label{fig:llm_size}
    \vspace{-0.5em}
\end{figure}

\begin{figure}[t]
    \centering
    \includegraphics[width=0.9\linewidth]{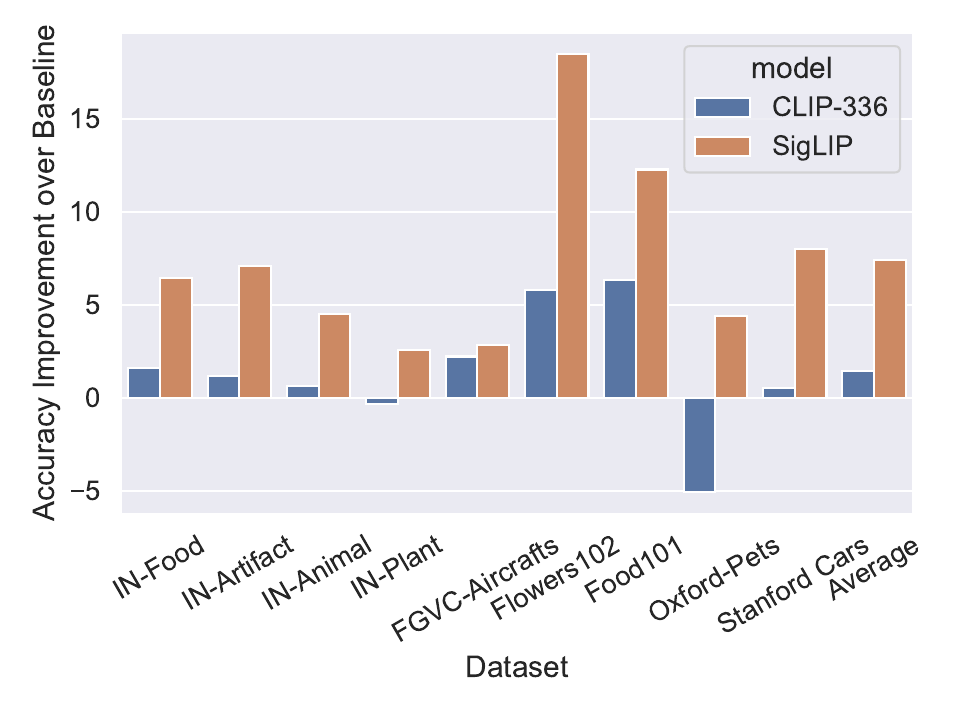}
    \caption{Improvements over our baseline when changing the \texttt{OpenAI ViT-L/14-224} image encoder to a higher resolution (336) or to \texttt{SigLIP SO400-224}.}
    \label{fig:image_encoder}
    \vspace{-1em}
\end{figure}

\begin{table}[t]
    \centering
     \def\arraystretch{0.97}
     \resizebox{1\linewidth}{!}{
    \begin{tabular}{lrrrr}
    \toprule
    \bf Model     &  \bf  IN-1k & \bf Train Half & \bf Test Half & \bf  \bench \\
         \midrule
\texttt{Baseline}    & 53.12 & 53.71 & 52.52 & 41.19 \\
\texttt{No Pretrain} & 51.94 & 51.56 & 52.32 & 38.71 \\
\texttt{Synthetic}  & 54.46	& 55.12	& 53.80 & 41.48 \\
\texttt{Template}   & 54.81	& 58.82	& 50.80 & 40.69 \\
\texttt{QA Task}    & 57.40	& 59.89	& 54.91 & 43.64 \\
\bottomrule
    \end{tabular}
    }
    \caption{Results for experiments with changes to the training data on: ImageNet-1k overall (IN-1k) and broken down for the training half and the held-out test half, and the average results over the 9 \bench datasets.}
    \label{tab:imagenet500}
    \vspace{-1em}
\end{table}

\rparagraph{LLM Size}
Larger LLMs generally make for better LVLMs, yielding better benchmark performance due to (\textit{inter alia}) improved reasoning capabilities \cite{liu_improved_2023,karamcheti_prismatic_2024,chu_mobilevlm_2024}.
Our multiple-choice object classification is not difficult from a reasoning or language-understanding perspective, but it requires familiarity with thousands of objects, which may be beyond the knowledge stored in smaller LLMs. For this analysis, we turn to the MobileVLM v2 model series \cite{chu_mobilevlm_2024}: with models trained on top of 1.7B, 3B, and 7B LLM backbones and otherwise identical architecture (image encoder and alignment module) and training procedure (data and training protocol for both the LLMs and subsequent LVLMs), we can isolate the effect of LLM size.
Figure~\ref{fig:llm_size} summarizes the results. Expectedly, the performance on all \bench datasets consistently improves with increased LLM size: we believe that this is because smaller LLMs simply encode less world knowledge and have semantically poorer representations for (fine-grained) objects.

\rparagraph{Image Encoder}
Following the observation that the quality of the CLIP image encoder may cap the LVLMs' performance (Figure~\ref{fig:plot:clip_lvlm_scatter}), we investigate the effect that LVLM's image encoder has on fine-grained object recognition. 
Our ``baseline'' LVLM aligns the \texttt{OpenAI CLIP-L/14-224} (CLIP-224 for short) image encoder with the 
LLM. We then create two other LVLMs by changing the image encoder with: (1) \texttt{OpenAI CLIP-L/14-336} (CLIP-336 for short), which takes images of \textit{larger resolution}  and (2) \texttt{SigLIP SO400M-224} (SigLIP for short) \cite{zhai_sigmoid_2023-1} as a `better' image encoder, boasting substantially higher benchmark results on image processing benchmarks. Figure~\ref{fig:image_encoder} summarizes the results. On one hand, encoding images in higher resolution (with CLIP-336, i.e., increasing from 224px to 336px) leads to only a marginal $\sim$1 accuracy point gain, averaged over all \bench datasets. The effect seems to depend on the object type: we see gains of over 5 points on Flowers102 \& Food102 but also a 5-point drop on Oxford-Pets.
The SigLIP encoder, on the other hand, greatly improves the baseline performance across the board.
The absolute gains of the SigLIP-based LVLM over the baseline LVLM (CLIP-224 encoder) are, however, not proportionate to gains that the corresponding SigLIP CLIP model yields over CLIP-224 in zero-shot object classification. For example, while SigLIP beats CLIP-224 by 27\% on FGVC-Aircraft,\footnote{Taken from: \href{https://github.com/mlfoundations/open_clip/blob/main/docs/openclip_classification_results.csv}{openclip\_classification\_results.csv}} the SigLIP-based LVLM beats the CLIP-224-based LVLM on the same dataset by only 3\%; inversely, on Food101, SigLIP has only a 2\% edge in CLIP comparison, but yields 12\% better performance in LVLM comparison. 

\rparagraph{Training Data}
The two LVLMs trained with most data, Idefics-2 and Qwen-VL (>1.5B images in total over both training stages) demonstrated the best performance on \bench (Table \ref{tbl:icbench}). 
As this scale of training is beyond the (computational) budget of most practitioners, we set to quantify the \bench gains from adding training data at smaller data scales, concretely at the data budget of LLaVA 1.5 (ca.\,1.2M images in total, see Table \ref{tab:model_infos}).    


\noindent\textit{Changes to Pretraining.} We hypothesize that a larger pretraining corpus benefits the LVLM due to having more of the objects named explicitly in the corresponding captions. We test this explicitly by replacing 25\% of the LLaVA 560k pretrainng images (with captions) with images from the ImageNet-1k train split. To have a held-out control set, we only use 500 of the 1000 classes (choosing every other class) for training; we select 280 images per class (140k training examples in total). 
We consider three training strategies for the added ImageNet images: \textbf{i)} with \textit{synthetic captions}, generated using BLIP \cite{li_blip_2022-1} (\texttt{Synthetic}); this setup tests the effect of images with objects but with captions that do not necessarily name them (e.g., for an image of a Keeshond, BLIP-generated caption will likely contain `\textit{dog}' but not `\textit{Keeshond}'). \textbf{ii)} with  \textit{template captions} (\texttt{Template}) such as \textit{``a picture of a \$label.''}; such captions are not visually descriptive but explicitly name the object in the image. \textbf{iii)} we skip the pretraining phase entirely (\texttt{No Pretrain}) and perform the task mix training on the randomly initialized alignment module; on standard benchmarks, skipping pretraining has been reported not to notably affect performance \cite{karamcheti_prismatic_2024}.

\noindent\textit{Changes to Task Mix Phase.} We incorporate ImageNet as an open-ended \texttt{QA Task} where the model is prompted to name the image object without candidate answers. We use the open-ended QA formulation in training to avoid model adaptation to the multiple-choice formulation of the task we use at test time on \bench. We again use 500 (out of the 1000) ImageNet classes and sample 150 examples per class (75k training examples in total). We do not otherwise change the LLaVA task mix data.

\noindent{\textit{Results.}} We report the results of this ablation in Table~\ref{tab:imagenet500}.  
Skipping the pretraining step entirely (\texttt{No Pretrain}) reduces the average \bench performance by over 2 accuracy points: this suggest that pretraining of the alignment module on image-text pairs is important for fine-grained object classification, unlike what was recently reported for other tasks \cite{karamcheti_prismatic_2024}. 
Training on images with both \texttt{Synthetic} and \texttt{Template} captions has a very limited effect on \bench performance and the unseen Test Half of ImageNet. Training on \texttt{Synthetic} brings a $\sim 1.5$-point gain for the 500 ImageNet object classes seen in training (Train Half in Table \ref{tab:imagenet500}); in comparison, the Template captions  bring a much more significant gain of 5\% for seen object classes: 
this strongly suggests that \textit{explicitly mentioning} the objects in the captions is key for learning the alignment module that allows LVLMs better fine-grained object classification; just having images containing the object does not suffice (or is, at least, less effective). Note that only the feed-forward alignment module is trained in the first phase, so the improvements with \texttt{Template} captions can only be the result of having learned a better alignment and not due to the image encoder or LLM (both frozen) obtaining better representations of objects and their mentions, respectively.
Including ImageNet as open-ended \texttt{QA Task} to the second task mix training phase has a larger effect on performance. For 500 of ImageNet-1k seen in training (Train Half), we observe a 6\% improvement, but also a 2-point improvement on the images from the held-out Test Half and on \bench.



\section{Conclusion}

In this work, we evaluate the capabilities of LVLMs for fine-grained object classification over different domains. We address the ambiguity of open-ended QA-based object classification evaluation and propose to replace it with a multiple-choice formulation, in which we retain the task difficulty by mining difficult (semantically closest classes) choices with a CLIP model. This way, we create \bench, a novel benchmark consisting of 9 fine-grained multi-choice object classification datasets. 
%
We benchmark 12 public LVLMs, demonstrating that their performance on \bench is largely uncorrelated with that on other image understanding and reasoning benchmarks: this renders fine-grained object classification a skill that is complementary to what the existing benchmarks test the LVLMs for. Our ablations identify the quality of the image encoder and the amount of explicit caption mentions of image objects in LVLM training data as factors that drive the performance.  
%
We hope our work stimulates wider research efforts on improving LVLMs for fine-grained object classification, in particular conceptual innovation (e.g., more effective training data and protocols for object classification with LVLMs) that goes well beyond mere scaling of LVLM pretraining to billions of image-text pairs.


\newpage

\section*{Limitations}

We identify three main limitations for our work:

First, while the goal of this work is not to evaluate every possible domain, we still likely exhibit a bias towards Anglospheric concepts as multiple datasets were created at British and US universities and use images sourced from the English internet.
ImageNet in particular shows such biases \cite{liu_visually_2021-1} in image source and for its classes.
While we briefly consider performance over geographic distribution shifts in the Appendix, we still likely overestimate performance for diverse cultural objects and concepts from around the globe.

Another limitation stems from the multiple-choice formulation: while it allows for well-defined answers, users `in the wild' are more likely to use an open-ended formulation.
While we expect results between the two formulations to correlate, some objects may be harder to classify in a multiple-choice setup due to the presence of challenging confounder options, and vice versa, some objects may be easier to classify in multiple-choice with the correct name as an option.

Finally, we only evaluate public LVLMs using LLMs of 7B parameters or less.
We do not consider larger models (e.g., LLaVA 1.5 with Vicuna-13B) or proprietary LVLMs (e.g., GPT4 \cite{openai_gpt-4_2023-1} or Gemini \cite{anil_gemini_2023}) because the inference time is too high on our compute (or not possible at all VRAM-wise) for the former and too expensive with >100,000 of API calls for the latter.

\section*{Acknowledgements}

This work was in part supported by the Alexander von Humboldt Foundation.

\bibliography{custom} 
\bibliographystyle{acl_natbib}

\appendix
\lstdefinestyle{mystyle}{
    basicstyle=\ttfamily\footnotesize,
    backgroundcolor=\color{lightgray},
    frame=single,
    breaklines=true,
    numbers=left,
    numberstyle=\tiny\color{gray},
    escapeinside={*@}{@*}
}
\lstset{style=mystyle}
\newcommand{\italic}[1]{\textit{#1}}

\begin{table}[t]
    \centering
     \def\arraystretch{0.97}
     \resizebox{1\linewidth}{!}{
    \begin{tabular}{ll}
    \toprule
    \bf Model & \bf Checkpoint \\
    \midrule
    Idefics-1 \cite{laurencon_obelisc_2023} & HuggingFaceM4/idefics-9b-instruct \\
    Idefics-2 \cite{laurencon_what_2024} & HuggingFaceM4/idefics2-8b \\
    BLIP2 Flan-T5-XL \cite{li_blip-2_2023} & Salesforce/blip2-flan-t5-xl \\
    InstructBLIP Flan-T5-XL \cite{dai_instructblip_2023} & Salesforce/instructblip-flan-t5-xl \\
    InstructBLIP Vicuna \cite{dai_instructblip_2023} & Salesforce/instructblip-vicuna-7b \\
    InternLM XComposer 2 \cite{dong_internlm-xcomposer2_2024} & internlm/internlm-xcomposer2-vl-7b \\
    LLaVA 1.5 \cite{liu_improved_2023} & llava-hf/llava-1.5-7b-hf \\
    LLaVA-Next (Mistral) \cite{liu_llava-next_2024} & llava-hf/llava-v1.6-mistral-7b-hf \\
    MobileVLM V2 \cite{chu_mobilevlm_2024} & mtgv/MobileVLM\_V2-7B \\
    Pali-Gemma $^1$ & google/paligemma-3b-mix-224 \\
    Phi-3-Vision \cite{abdin_phi-3_2024} & microsoft/Phi-3-vision-128k-instruct \\
    Qwen-VL-Chat \cite{bai_qwen-vl_2023} & Qwen/Qwen-VL-Chat \\
    \bottomrule
    \end{tabular}
    }
    \caption{The tested public LVLM with the corresponding checkpoint from HuggingFace we used.
    $^1$\href{https://ai.google.dev/gemma/docs/paligemma/model-card}{Model Card}, tech report pending at time of writing.
    }
    \label{tab:appendix:model_infos}
\end{table}

\section{Evaluation Details}
\label{sec:appendix:eval_details}

\paragraph{Models \& Inference:}
In Table~\ref{tab:appendix:model_infos}, we specify the exact checkpoint we used for each model. 
We adapt the respective official code of each model for inference.
All models use greedy decoding.

We use the following prompt for all models. Depending on the task, we change the default question at the beginning to prime the model for the dataset domain:
\begin{lstlisting}
*@\textbf{\textrm{Default:}}@* Which of these choices is shown in the image?
*@\textbf{\textrm{IN-Animal:}}@* Which of these animals is shown in the image?
*@\textbf{\textrm{IN-Plant:}}@* Which of these plants is shown in the image?
*@\textbf{\textrm{FGVC-Aircraft:}}@* Which of these aircrafts is shown in the image?
*@\textbf{\textrm{Flowers102:}}@* Which of these flowers is shown in the image?
*@\textbf{\textrm{Food101:}}@* Which of these dishes is shown in the image?
*@\textbf{\textrm{Oxford-Pet:}}@* Which of these pets is shown in the image?
*@\textbf{\textrm{Stanford-Cars:}}@* Which of these cars is shown in the image?
Choices:
A. $CHOICE1
B. $CHOICE2
C. $CHOICE2
D. $CHOICE3
Answer with the letter from the given choices directly.
\end{lstlisting}
We expect the model to answer with a letter and count the example as correct if the generated answer begins with the letter corresponding to the correct answer.

\paragraph{Dataset Details:}
In general, we evaluate on the full test split (or, if no public test split exists like with ImageNet, the validation split) of every dataset.

The datasets that we constructed from ImageNet-21k (Animal, Plant, Food, Artifact) are the exception: due to the large amount of classes, we only use 10 images per class instead of the full 50 to keep computation time manageable.
In addition, we use the processed version of ImageNet-21k and not the original (>1TB large) version for disk space reasons; the processed version has all images resized to 224$\times$224px.
During creating of the four datasets, we remove all classes that have no unique label (keeping only the first occurrence of a label) to achieve a 1-to-1 mapping between classes and labels.

\section{Training Details}
\label{sec:appendix:train_details}

We closely follow the architecture and training protocol of LLaVA 1.5 \cite{liu_improved_2023}.
As LLM, we use the instruction-trained StableLM 2 1.6B Zephir \cite{bellagente_stable_2024} (\texttt{stabilityai/stablelm-2-zephyr-1\_6b}), which is a small but performant LLM.
The default image encoder is \texttt{OpenAI CLIP ViT-L/14-224}.
Training is done on a single NVIDIA RTX 3090 with training one model taking less than 2 days.

We train the models using AdamW optimizer \cite{loshchilov_decoupled_2019} with a cosine learning rate decay schedule.
For the pre-training phase, we use learning rate 1e-3, weight decay 0, and batch size 256.
For the task-mix training phase, we use learning rate 2e-4, weight decay 0, and batch size 128; we do not fine-tune the full LLM but apply LoRA \cite{hu_lora_2022-2} to all weights with $r=64$, $\alpha=128$.

\section{LVLM Performance on Popular Benchmarks}
\label{sec:appendix:general_eval}
We collate public results on select popular benchmarks for evaluating LVLMs (GQA \cite{hudson_gqa_2019}, MMBench \cite{liu_mmbench_2023}, and MMMU \cite{yue_mmmu_2023}) for the models of Table~\ref{tab:model_infos}. 
Comparing these results against the performance in object classification shows that the latter is an independent skill that does not directly correlate with these benchmarks.

\begin{table*}[t]
    \centering
    \begin{tabular}{lrrr}
    \toprule
    \bf Model & \bf GQA & \bf MMBench & \bf  MMMU \\
    \midrule
    Idefics-1 & --- & 35.2 & 28.7 \\
    Idefics-2 & --- & 76.8 & 43.5 \\
    BLIP2 Flan-T5-XL & *44.0 & --- & 34.4 \\
    InstructBLIP Flan-T5-XL & *48.4 & --- & 32.9 \\
    InstructBLIP Vicuna 7B & *49.2 & 38.3 & --- \\
    InternLM XComposer 2 & --- & 79.6 & 43.0 \\
    LLaVA 1.5 7B & 62.0 & 64.3 & --- \\
    LLaVA-Next Mistral 7B & 64.8 & 68.7 & --- \\
    MobileVLM V2 7B & 62.6 & 69.2 & --- \\
    Pali-Gemma & **65.6 & --- & --- \\
    Phi-3-Vision & --- & 80.5 & 40.4 \\
    Qwen-VL-Chat & 57.5 & 60.6 & 35.9 \\
    \bottomrule
    \end{tabular}
    \caption{Performance on standard benchmarks for image understanding and reasoning.
    * unlike other models, has not included GQA in training task mix.
    ** with model fine-tuned on GQA, not the mix version used for testing.
    }
    \label{tab:appendix:standard_benchmarks}
\end{table*}

\section{Additional Evaluation on More Datasets}
\label{sec:appendix:more_eval}
In this section, we consider general object classification datasets (not covering a specific domain) and consider how LVLMs handle image distribution shifts for the same object using ImageNet and its variants and GeoDE \cite{ramaswamy_geode_2023}.

\paragraph{ImageNet Image Distribution Shifts.}
\label{sec:appendix:more_eval:imagenet}

There are several datasets that collect new images for the classes of ImageNet-1k \cite{deng_imagenet_2009}, or at least for a subset of them.
Here, we consider \textbf{ImageNet-Adversarial} \cite{hendrycks_natural_2021}, which contains images for 200 classes that are difficult to correctly classify for a model trained on the ImageNet-1k training split; \textbf{ImageNet-Rendition} \cite{hendrycks_many_2021}, which contains for 200 classes images of the objects where the image is painted, a plushy, origami, or other renditions; and \textbf{ImageNet-Sketch} \cite{wang_learning_2019-1}, which contains black-and-white drawings for all 1000 classes.

CLIP models generally excel at transferring between the different image distributions due to their large-scale training \cite{radford_learning_2021-1}.
We evaluate in Table~\ref{tab:appendix:imagenet} if LVLMs see similar results despite training the alignment with the image encoder on magnitudes less data and generally only with natural images.
We observe that the ranking between the models is similar to our evaluation on \bench in Table~\ref{tbl:icbench}.
The changes in accuracy from ImageNet-1k to the variants are qualitatively similar to the underlying CLIP models for the LVLMs.
This suggests that other representations of objects (like sketches) are encoded similarly enough by the image encoder that the LVLM can `recognize' without extra training on different image types.

\begin{table*}[t]
\centering
\begin{tabular}{lrrrrr}
\toprule
    \bf models & \bf IN-1k & \bf IN-adversarial & \bf IN-rendition & \bf IN-sketch & \bf $\varnothing$ \\
\midrule 
Idefics-1  & 60.09 & 50.03 & 72.20 & 50.13 & 58.11 \\
Idefics-2  & 73.39 & 79.84 & 93.23 & 68.21 & 78.67 \\
BLIP-2 Flan-T5-XL  & 66.12 & 67.48 & 90.48 & 64.85 & 72.23 \\
InstructBLIP Flan-T5-XL  & 66.15 & 69.69 & 90.58 & 64.46 & 72.72 \\
InstructBLIP Vicuna  & 56.27 & 59.75 & 76.82 & 54.84 & 61.92 \\
InternLM XComposer 2  & 65.65 & 73.08 & 83.29 & 56.99 & 69.75 \\
LLaVA 1.5  & 62.44 & 68.53 & 79.30 & 55.88 & 66.54 \\
LLaVA-Next  & 60.86 & 67.20 & 78.12 & 53.50 & 64.92 \\
MobileVLM v2  & 61.16 & 64.59 & 79.63 & 54.66 & 65.01 \\
Pali-Gemma  & 69.56 & 68.45 & 92.15 & 65.55 & 73.93 \\
Phi-3-Vision  & 61.71 & 56.71 & 79.18 & 56.01 & 63.40 \\
Qwen-VL-Chat  & 71.20 & 70.99 & 90.59 & 67.16 & 74.98 \\

\bottomrule
\end{tabular}
\caption{Results for ImageNet-1k and four distribution-shifted versions.}
\label{tab:appendix:imagenet}
\end{table*}

\paragraph{Geographic Shifts with GeoDE.}
\label{sec:appendix:more_eval:geode}

We now consider geographic distribution shift using GeoDE \cite{ramaswamy_geode_2023}, a dataset with 40 classes for which there are images evenly distributed around the globe for six regions: Europe, Africa, Southeast Asia, West Asia, East Asia, and the Americas (which does not include here the US or Canada).
Results of the tested LVLMs are reported in Table~\ref{tab:appendix:geo}.
While GeoDE is a generally easy dataset with high accuracy throughout, we still observe substantial differences between the regions:
European images consistently enjoy the highest accuracy, all non-African regions follow close by with 0-3 points worse than Europe, and finally, the African images noticeably trail behind by 2-4 points lower accuracy compared to the overall average accuracy.
This shows that geographic biases in the training data, both for the image encoder and for the LVLM \cite{pouget2024filter}, result in disadvantages for large parts of the population. 


\begin{table*}[t]
    \centering
     \def\arraystretch{0.97}
     \resizebox{1\linewidth}{!}{
    \begin{tabular}{lrrrrrrr}
    \toprule
\bf models & \bf Europe & \bf Africa & \bf Southeast Asia & \bf Americas & \bf West Asia & \bf East Asia &  \bf All \\
\midrule 
Idefics-1  & 85.48 & 79.85 & 84.65 & 83.61 & 84.06 & 84.22 & 83.56 \\
Idefics-2  & 90.15 & 86.59 & 90.00 & 89.40 & 90.03 & 89.65 & 89.23 \\
BLIP-2 Flan-T5-XL  & 91.24 & 87.49 & 90.91 & 89.64 & 90.45 & 89.32 & 89.79 \\
InstructBLIP Flan-T5-XL  & 88.64 & 84.10 & 88.46 & 86.87 & 87.99 & 87.60 & 87.20 \\
InstructBLIP Vicuna  & 76.36 & 70.53 & 75.61 & 75.34 & 74.78 & 76.19 & 74.70 \\
InternLM XComposer 2  & 91.54 & 87.59 & 90.81 & 90.48 & 90.63 & 89.54 & 90.04 \\
LLaVA 1.5  & 86.06 & 82.81 & 86.27 & 84.66 & 86.35 & 84.17 & 84.99 \\
LLaVA-Next  & 86.75 & 82.90 & 86.55 & 85.35 & 85.46 & 84.65 & 85.24 \\
MobileVLM v2  & 82.13 & 75.16 & 79.27 & 79.26 & 81.09 & 77.99 & 79.05 \\
Pali-Gemma  & 90.94 & 87.12 & 90.53 & 90.14 & 90.68 & 90.09 & 89.84 \\
Phi-3-Vision  & 89.76 & 86.49 & 89.44 & 88.75 & 88.46 & 87.61 & 88.39 \\
Qwen-VL-Chat  & 90.94 & 87.31 & 88.86 & 89.24 & 90.79 & 89.32 & 89.34 \\
\bottomrule
    \end{tabular}
    }
    \caption{Result on the GeoDE dataset for each region and the overall accuracy for all examples together.}
    \label{tab:appendix:geo}
\end{table*}

\section{More CLIP Results}
\begin{figure*}[t]
    \centering
\begin{subfigure}[b]{0.3\textwidth}
    \includegraphics[width=0.99\linewidth]{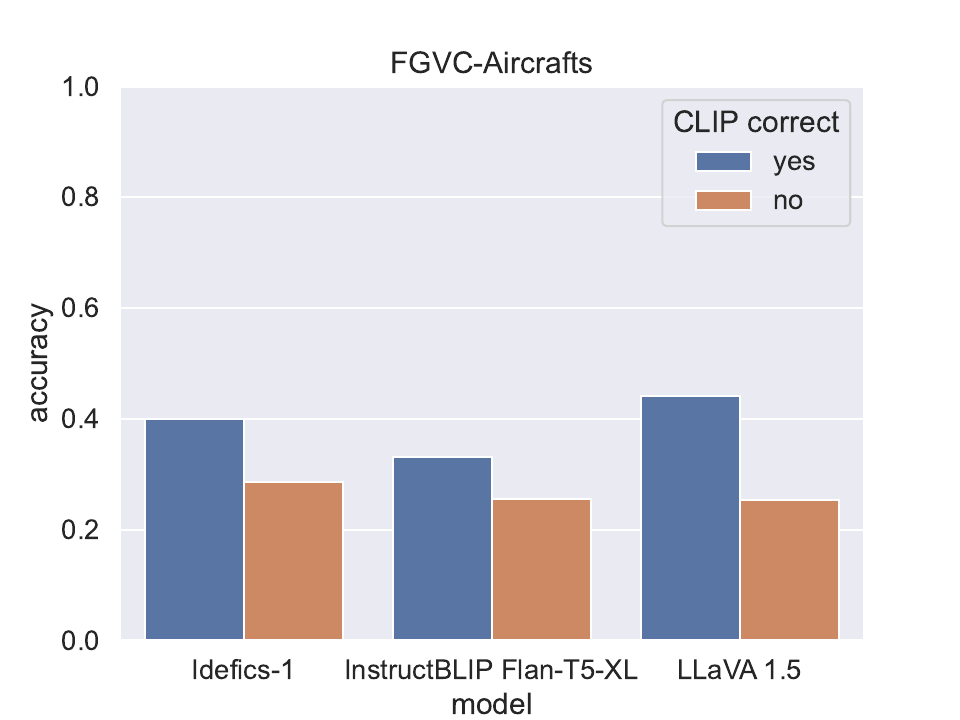}
\end{subfigure}
\begin{subfigure}[b]{0.3\textwidth}
    \includegraphics[width=0.99\linewidth]{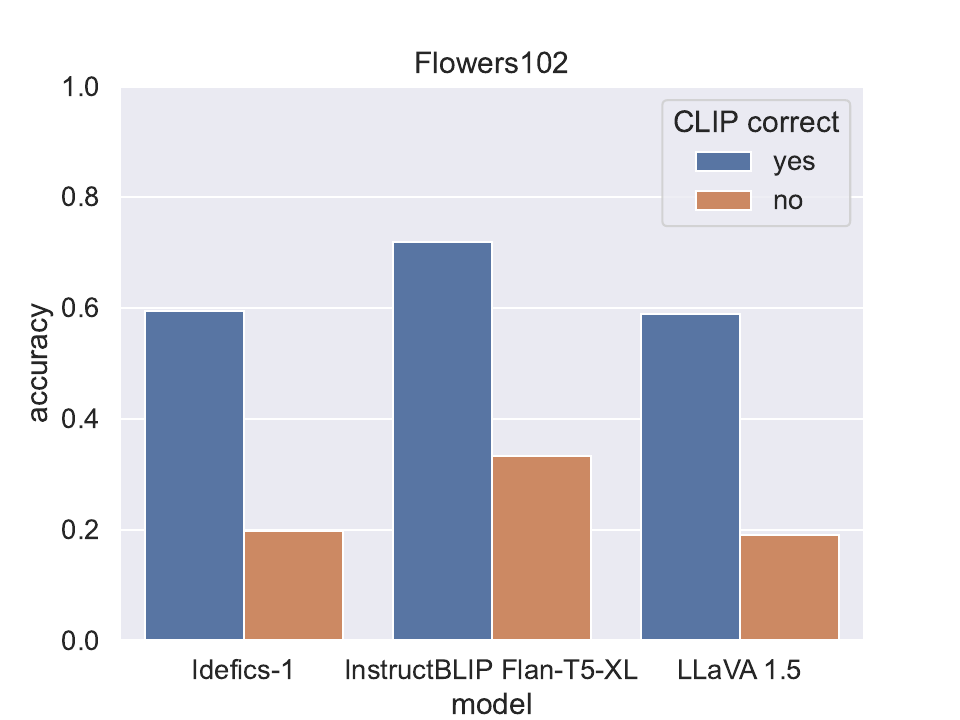}
\end{subfigure}
\begin{subfigure}[b]{0.3\textwidth}
    \includegraphics[width=0.99\linewidth]{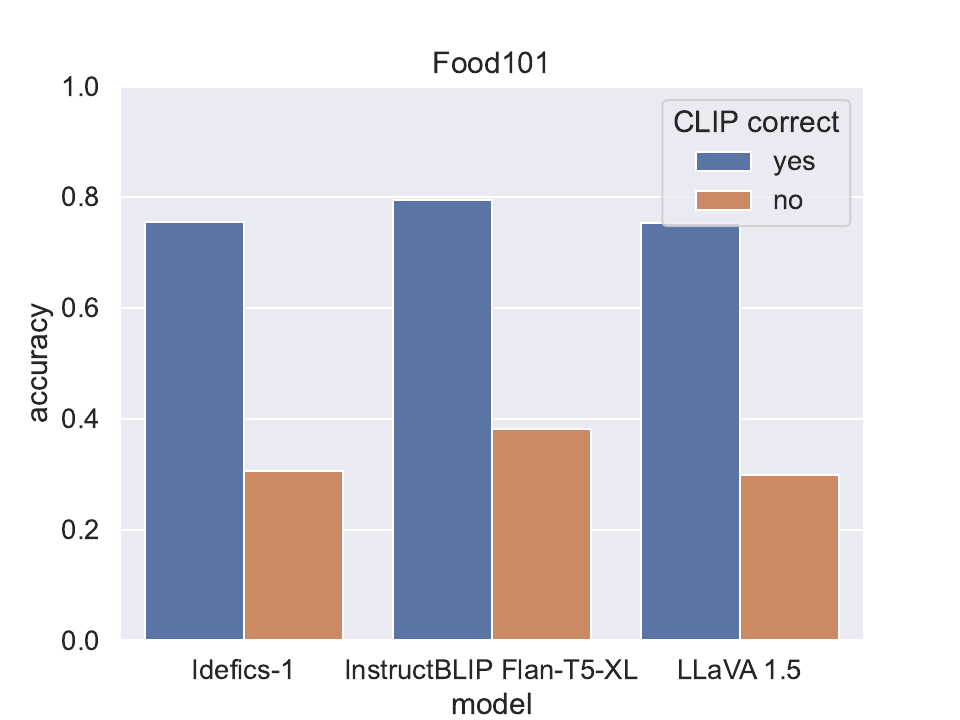}
\end{subfigure}
\begin{subfigure}[b]{0.3\textwidth}
    \includegraphics[width=0.99\linewidth]{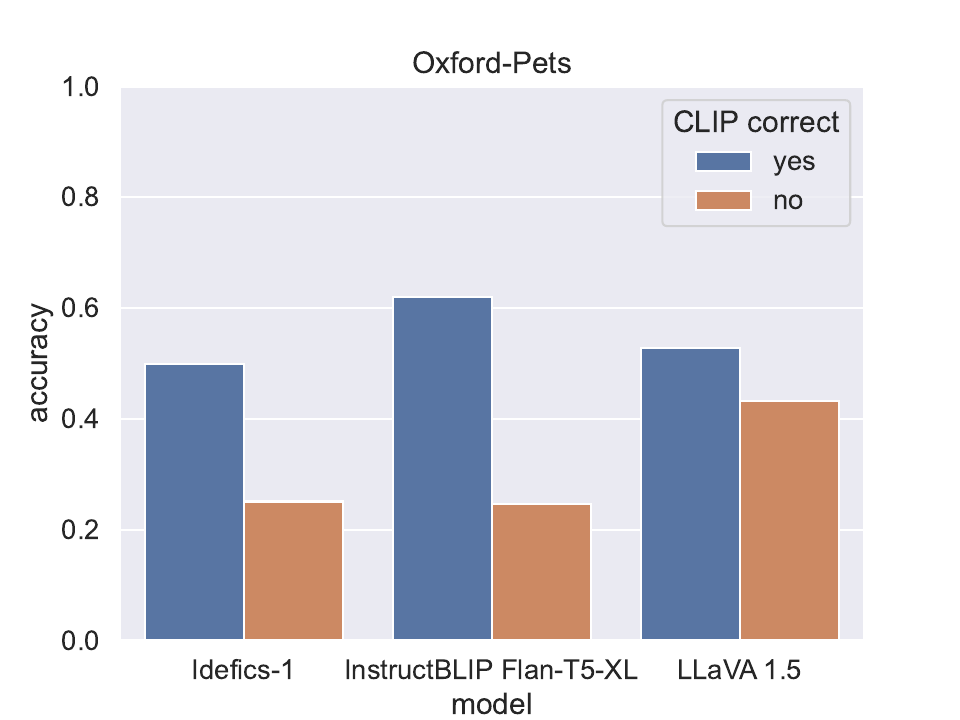}
\end{subfigure}
\begin{subfigure}[b]{0.3\textwidth}
    \includegraphics[width=0.99\linewidth]{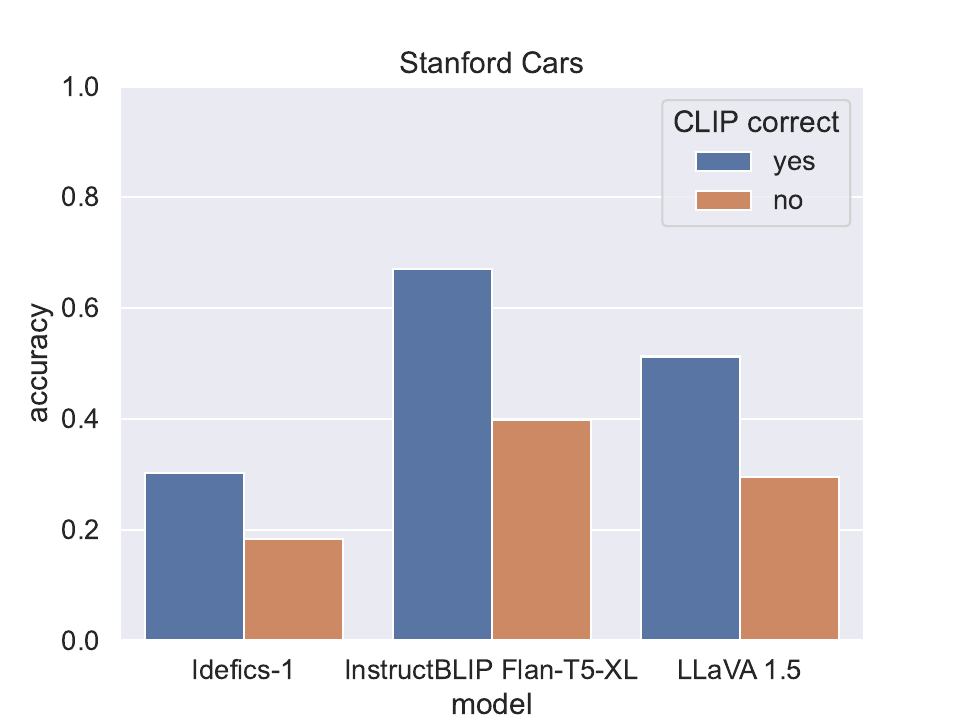}
\end{subfigure}
\begin{subfigure}[b]{0.3\textwidth}
    \includegraphics[width=0.99\linewidth]{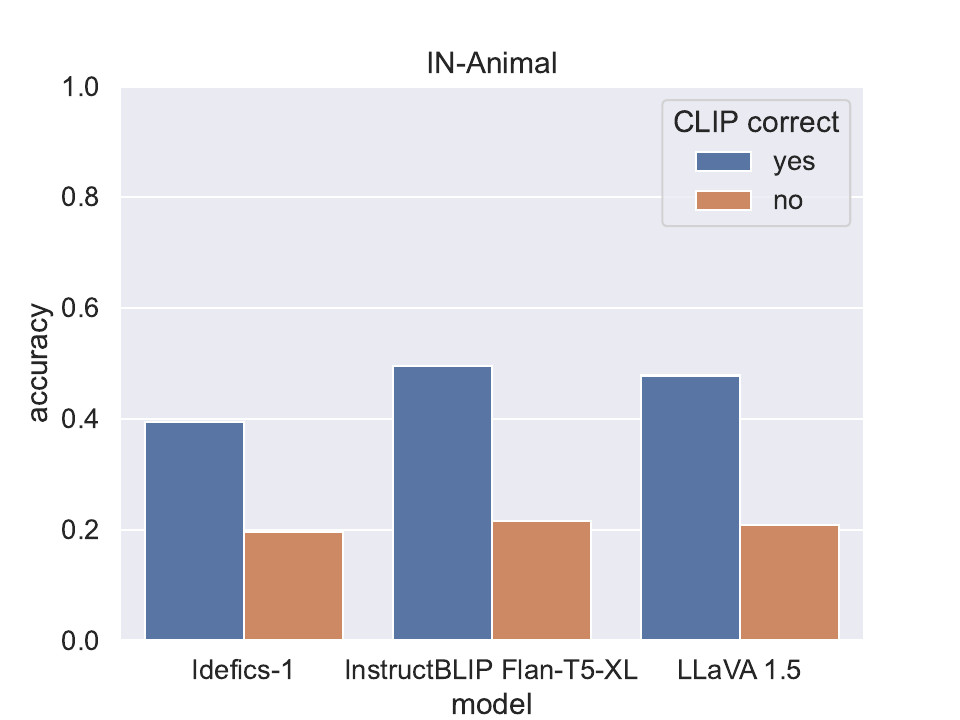}
\end{subfigure}
\begin{subfigure}[b]{0.3\textwidth}
    \includegraphics[width=0.99\linewidth]{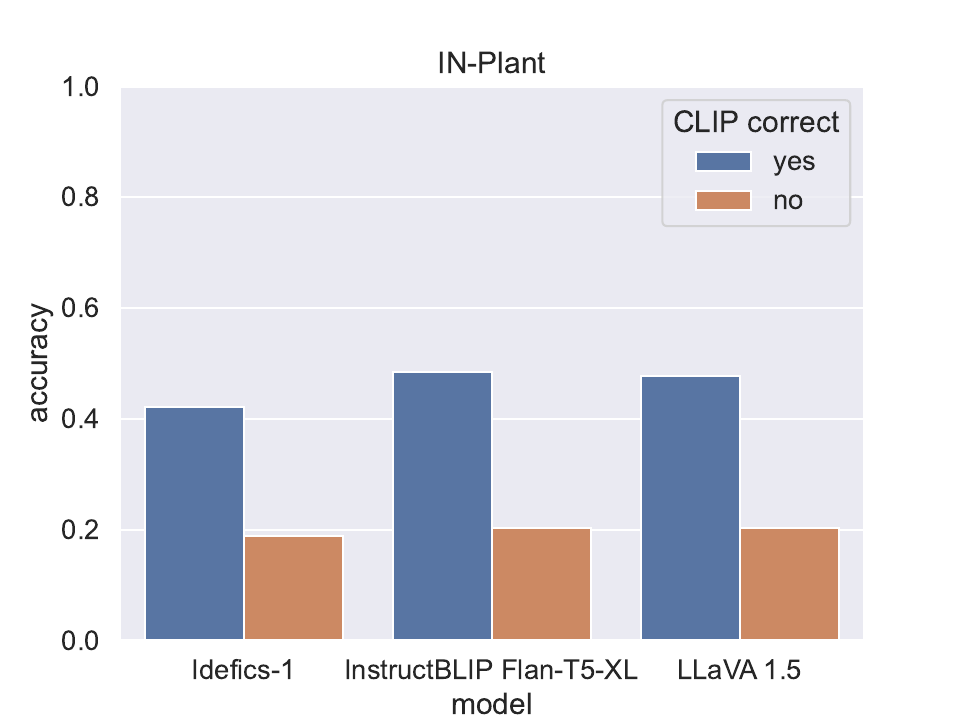}
\end{subfigure}
\begin{subfigure}[b]{0.3\textwidth}
    \includegraphics[width=0.99\linewidth]{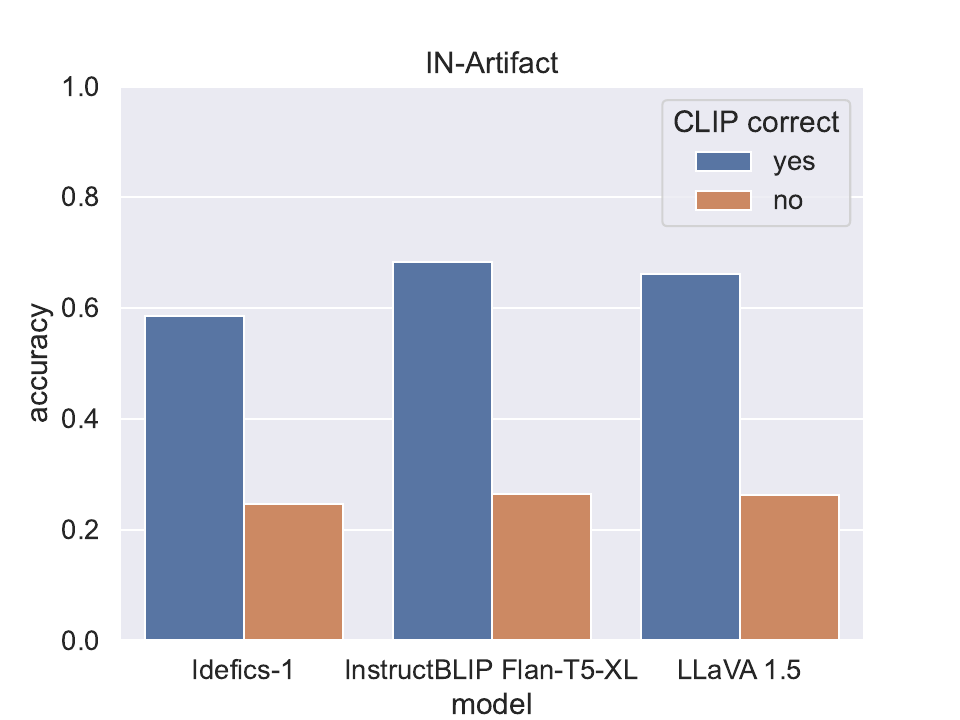}
\end{subfigure}
\begin{subfigure}[b]{0.3\textwidth}
    \includegraphics[width=0.99\linewidth]{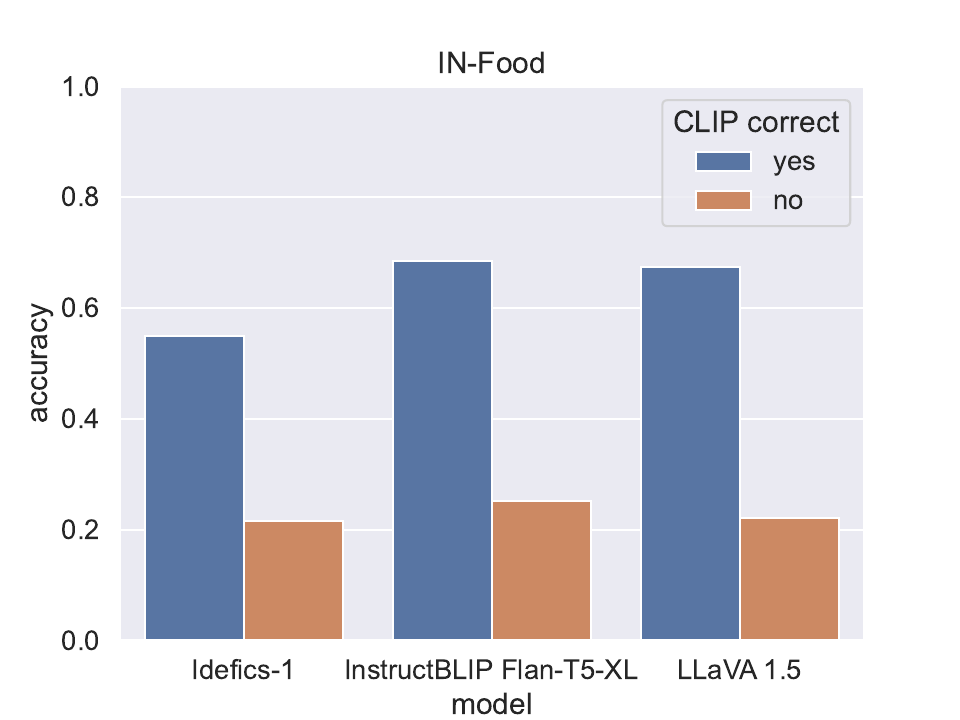}
\end{subfigure}
    \caption{Conditionally accuracy on different datasets of different models if the CLIP image encoder would (in)correctly classify an example in zero-shot.}
    \label{fig:appendix:clip_error}
\end{figure*}

We present the results for conditional accuracy of LVLMs for all datasets in Figure~\ref{fig:appendix:clip_error}.

\section{Full Experiment Results}

Complementary to the Figures in the main paper, we report the raw results of MobileVLMv2 in Table~\ref{tab:appendix:llm_size} and for our trained models in Table~\ref{tbl:appendix:experiments}.

\begin{table*}[t]
    \centering
     \def\arraystretch{0.97}
     \resizebox{1\linewidth}{!}{
    \begin{tabular}{lrrrrrrrrrr}
    \toprule
\bf Model & \bf IN-food & \bf IN-artifact & \bf IN-animal & \bf IN-plant & \bf Aircraft & \bf Flowers102 & \bf Food101 & \bf O.-Pet & \bf S.-Cars & \bf $\varnothing$ \\
\midrule 
1.7B  & 40.89 & 38.67 & 30.36 & 28.19 & 29.07 & 47.72 & 61.03 & 41.26 & 36.08 & 39.25 \\
3B  & 44.28 & 42.30 & 34.83 & 31.97 & 32.85 & 47.70 & 68.67 & 46.69 & 41.52 & 43.42 \\
7B  & 46.50 & 44.58 & 37.60 & 33.75 & 35.01 & 54.89 & 74.38 & 53.69 & 46.29 & 47.41 \\
\bottomrule
\end{tabular}
}
\caption{Results for the three sizes of MobileVLM v2.}
\label{tab:appendix:llm_size}
\end{table*}

\begin{table*}[t]
    \centering
     \def\arraystretch{0.97}
     \resizebox{1\linewidth}{!}{
    \begin{tabular}{lrrrrrrrrrr}
    \toprule
\bf models & \bf IN-food & \bf IN-artifact & \bf IN-animal & \bf IN-plant & \bf fgvc aircraft & \bf flowers102 & \bf food101 & \bf oxford pet & \bf stanford cars & \bf $\varnothing$ \\
\midrule
\texttt{Baseline}  & 43.43 & 40.33 & 32.18 & 31.54 & 30.27 & 38.33 & 62.40 & 50.12 & 42.08 & 41.19 \\
\texttt{CLIP-336}  & 45.01 & 41.51 & 32.81 & 31.20 & 32.49 & 44.15 & 68.73 & 45.05 & 42.61 & 42.62 \\
\texttt{SigLIP}  & 49.88 & 47.44 & 36.70 & 34.11 & 33.09 & 56.82 & 74.69 & 54.51 & 50.08 & 48.59 \\
\texttt{No Pretrain}  & 41.55 & 39.63 & 31.50 & 29.80 & 30.30 & 40.30 & 58.44 & 40.23 & 36.60 & 38.71 \\
\texttt{Synthetic}   & 43.69 & 40.94 & 32.85 & 31.04 & 32.16 & 39.68 & 64.61 & 47.48 & 40.90 & 41.48 \\
\texttt{Template}    & 44.74 & 39.93 & 32.79 & 31.09 & 30.09 & 38.07 & 62.80 & 46.31 & 40.43 & 40.69 \\
\texttt{QA Task}       & 44.00 & 41.45 & 33.77 & 31.55 & 32.34 & 49.16 & 67.63 & 51.62 & 41.24 & 43.64 \\
    \bottomrule
    \end{tabular}
    }
    \caption{
    Full results for our trained models.
    }
    \label{tbl:appendix:experiments}
\end{table*}

\end{document}